\documentclass[11pt,a4paper]{article}

\usepackage[margin=1in]{geometry}
\usepackage{setspace}
\usepackage{microtype}
\usepackage[absolute,overlay]{textpos}

\usepackage{amsmath,amssymb,amsfonts}
\usepackage{booktabs}
\usepackage{array}
\usepackage{multirow}

\usepackage{graphicx}
\usepackage{float}
\usepackage{caption}
\usepackage{subcaption}

\usepackage{enumitem}
\usepackage[hidelinks]{hyperref}
\usepackage{tikz}
\usetikzlibrary{positioning, arrows.meta, shapes.geometric}

\setlist[itemize]{leftmargin=*, itemsep=0.25em}
\setlist[enumerate]{leftmargin=*, itemsep=0.25em}
\newcommand{\reserve}{V}
\newcommand{\rprice}{r_{\mathrm{price}}}
\newcommand{\rscen}{r_{\mathrm{scenario}}}
\newcommand{\premium}{P}
\newcommand{\sumassured}{S}
\newcommand{\mortality}{\mu}

\newcommand{\optionalfigure}[4]{
\begin{figure}[H]
\centering
\IfFileExists{#1}{
    \includegraphics[width=#2\textwidth]{#1}
}{
    \fbox{
    \begin{minipage}[c][0.20\textheight][c]{0.82\textwidth}
    \centering
    Figure placeholder: \texttt{#1}
    \end{minipage}}
}
\caption{#3}
\label{#4}
\end{figure}
}

\title{
Insurance Reserve Intelligence Platform\\[0.3cm]
\large A Physics- and Knowledge-Informed Neural Framework for Term-Life Reserve Prediction, Evaluation, Sensitivity Analysis, and Optimization
}

\author{
Anugya A*\\
\small anugyaa@iisc.ac.in
\and
Saket Mohanty\\
\small saket.mohanty@exlservice.com
\and
Abhilash Timmapur\\
\small abhilash.timmapur@exlservice.com
\and
Somya Rai\\
\small somya.rai@exlservice.com
}
\begin{document}

\maketitle

\begin{textblock*}{6cm}(2cm,27.75cm)
\footnotesize\textsuperscript{*}Primary author
\end{textblock*}

\begin{abstract}
Insurance reserve estimation is a fundamental actuarial task that supports premium pricing, solvency assessment, financial reporting, capital planning, and risk management. Classical reserve methods based on Thiele's differential equation provide a rigorous and interpretable foundation for life insurance valuation. However, repeated reserve valuation becomes increasingly expensive when reserve calculations are embedded within sensitivity analysis, optimization, or large-scale scenario evaluation.

This paper presents an Insurance Reserve Intelligence Platform for term-life reserve modelling using a classical Thiele-equation solver and a Physics-Informed Neural Network (PINN) enhanced with Knowledge-Informed Neural Network (KINN) losses. The framework includes synthetic policy generation, risk-adjusted premium calculation, classical reserve trajectory generation, reserve-ratio dataset construction, configurable neural training, validation diagnostics, sensitivity and elasticity analysis, prototype optimization workflows, and interest-rate scenario diagnostics. A major refinement in the project is the use of premium ratio and the explicit separation of pricing-time and scenario-time interest-rate semantics.

The final model uses a seven-dimensional feature representation consisting of elapsed time, issue age, pricing interest rate, scenario interest rate, premium ratio, sum assured, and mortality intensity. The model predicts a standardized reserve ratio rather than a raw monetary reserve, improving numerical stability across policies with different sums assured. The reported final model achieved an $R^2$ score of 0.9887, with MAE 785.48 and RMSE 1212.76 on the test dataset. A runtime benchmark on 200 policies showed that PINN/KINN inference was approximately 119.53 times faster than the classical solver. Validation results show strong aggregate prediction accuracy, strong physics residual behaviour, and good boundary performance, but also reveal remaining weaknesses in monotonicity and out-of-distribution generalization.
\end{abstract}

\section{Introduction}

Insurance reserves represent the amount an insurer must hold to meet future policyholder obligations. In life insurance, reserves directly influence pricing, solvency monitoring, financial reporting, regulatory compliance, capital management, and risk governance. Reserve estimation is therefore not only a numerical modelling task but also a core actuarial function.

In practice, actuaries require more than a single reserve estimate. They need reserve trajectories over time, comparisons across policies, sensitivity studies, and controlled experiments under changed assumptions. These use cases require repeated reserve calculations. Classical actuarial solvers are reliable and interpretable, but repeated numerical solving can be expensive when the number of policies, time points, scenarios, and optimization iterations increases.

This motivates the use of differentiable surrogate models. A neural reserve surrogate trained against a classical actuarial solver can approximate the reserve function and provide rapid reserve estimates after training. Physics-Informed Neural Networks (PINNs) are suitable for this problem because the governing actuarial equation can be included directly in the training objective. Knowledge-Informed Neural Networks (KINNs) extend this idea by incorporating actuarial constraints such as monotonicity, reserve ceilings, solvency behaviour, smoothness, and scenario-aware losses.

The objective of this project is to develop a research-grade Insurance Reserve Intelligence Platform for term-life insurance reserves. The platform combines synthetic policy generation, a classical Thiele solver, a neural reserve surrogate, a configurable loss framework, validation diagnostics, sensitivity analysis, and prototype optimization experiments. Broader production features such as real-data deployment, portfolio-scale governance, full stress-testing systems, and digital twin deployment are treated as future work.

\section{Project Scope}

This work focuses on the development of an Insurance Reserve Intelligence Platform for term-life insurance. The proposed framework integrates classical actuarial reserve computation with Physics-Informed Neural Networks and Knowledge-Informed Neural Networks to support reserve prediction, evaluation, sensitivity analysis, and prototype optimization.

The scope of the project includes:

\begin{itemize}
    \item Synthetic generation of term-life insurance policies using actuarial assumptions;
    \item Risk-adjusted premium calculation and mortality modelling;
    \item Reserve trajectory generation using Thiele's differential equation;
    \item Dataset preparation, feature normalization, and reserve-ratio target representation;
    \item Reserve prediction using a PINN/KINN model;
    \item A configurable data, physics-informed, knowledge-informed, and scenario-oriented loss functions;
    \item Model validation using prediction accuracy, PDE residuals, boundary conditions, monotonicity checks, and generalization metrics;
    \item Sensitivity and elasticity analysis for reserve interpretation;
    \item Prototype optimization workflows for pricing, product-design-style, and portfolio-style reserve experiments;
    \item Visualization and reporting tools for interpreting model performance and reserve behaviour.
\end{itemize}

The framework is modular, but this paper does not claim production deployment, real-data validation, complete baseline benchmarking, or a completed insurance digital twin. These are discussed as future work.

\section{Contributions}

The major contributions of this work are as follows.

\begin{enumerate}
    \item \textbf{Hybrid actuarial and neural reserve framework.}
    A classical Thiele-equation reserve solver is combined with a neural reserve surrogate. The solver provides benchmark reserve trajectories, while the neural model learns a differentiable approximation of the reserve surface.

    \item \textbf{Explicit PINN formulation.}
    The model is a PINN because the training objective includes the residual of Thiele's differential equation. The network is not trained only to fit reserve data; it is also penalized when its learned reserve surface violates the actuarial differential equation.

    \item \textbf{Explicit KINN formulation.}
    The model is a KINN because the training objective includes actuarial knowledge beyond the PDE, including boundary behaviour, monotonicity expectations, reserve ceiling constraints, solvency constraints, smoothness, and scenario-oriented losses.

    \item \textbf{Synthetic actuarial data generation.}
    The project develops synthetic term-life policy generation using issue age, term, interest rate, mortality, premium, sum assured, and underwriting assumptions.

    \item \textbf{Standardized reserve-ratio formulation.}
    The model predicts a standardized reserve ratio, $V/S$, rather than raw monetary reserve values. This improves numerical stability across heterogeneous policy sizes.

    \item \textbf{Premium-ratio representation.}
    Raw premium is replaced with premium ratio, $P/S$, so that premium and sum assured can be treated more independently during sensitivity analysis.

    \item \textbf{Evaluation, sensitivity, and prototype optimization framework.}
    The platform includes regression metrics, validation dashboards, finite-difference sensitivity analysis, elasticity analysis, reserve trajectory comparisons, runtime benchmarking, and optimization convergence plots.
\end{enumerate}

\section{Actuarial Background}

\subsection{Insurance Reserves}

A life insurance reserve is the financial liability held by an insurer to meet future policy obligations. For a term-life policy, reserves evolve during the policy term and return to zero at maturity if no claim occurs. Reserve values depend on elapsed time, age, mortality, interest rate, premium, and sum assured.

Accurate reserve estimation is essential because under-reserving can create solvency risk, while over-reserving can reduce capital efficiency. A reserve model should therefore be accurate, stable, and actuarially interpretable.

\subsection{Thiele's Differential Equation}

For continuous-time term-life insurance, reserve evolution can be described by Thiele's differential equation:
\begin{equation}
\frac{d\reserve(t)}{dt}
=
r\reserve(t)
+
\premium
-
\mortality(t)\left(\sumassured-\reserve(t)\right),
\label{eq:thiele}
\end{equation}
where $\reserve(t)$ is reserve, $r$ is the valuation interest rate, $\premium$ is premium, $\mortality(t)$ is mortality intensity, and $\sumassured$ is sum assured.

The terminal condition is:
\begin{equation}
\reserve(T)=0,
\label{eq:terminal}
\end{equation}
where $T$ is the policy term. The classical solver integrates the equation backward from maturity and provides the benchmark reserve trajectories used during training and evaluation.

\section{Dataset and Experimental Setup}

\subsection{Synthetic Dataset}

The project uses synthetic data because real insurance datasets were unavailable. The final configuration is summarized in Table~\ref{tab:dataset_statistics}.

\begin{table}[H]
\centering
\caption{Dataset and policy generation configuration.}
\label{tab:dataset_statistics}
\begin{tabular}{lc}
\toprule
Attribute & Value \\
\midrule
Training policies & 1000 \\
Validation policies & 200 \\
Test policies & 200 \\
Time steps per policy & 48 \\
Issue age range & 25 to 70 \\
Policy term range & 5 to 30 years \\
Interest-rate range & 1\% to 8\% \\
Sum assured range & 50,000 to 1,000,000 \\
Premium loading & 1.10 \\
Maximum expiry age & 80 \\
\bottomrule
\end{tabular}
\end{table}

The reserve statistics from the generated dataset are shown in Table~\ref{tab:reserve_statistics}.

\begin{table}[H]
\centering
\caption{Reserve distribution statistics from the generated dataset.}
\label{tab:reserve_statistics}
\begin{tabular}{lc}
\toprule
Statistic & Value \\
\midrule
Minimum reserve & -23,714.75 \\
Maximum reserve & 79,262.70 \\
Mean reserve & 6,445.48 \\
Standard deviation & 11,408.39 \\
\bottomrule
\end{tabular}
\end{table}

Exploratory data analysis was used to check whether the generated policies covered the intended actuarial space. Figure~\ref{fig:reserve_vs_age} shows the relationship between issue age and peak reserve in the synthetic term-life portfolio. The plot is used as an exploratory data analysis check to verify that the generated reserves vary meaningfully with actuarial risk factors and policy duration.

\optionalfigure{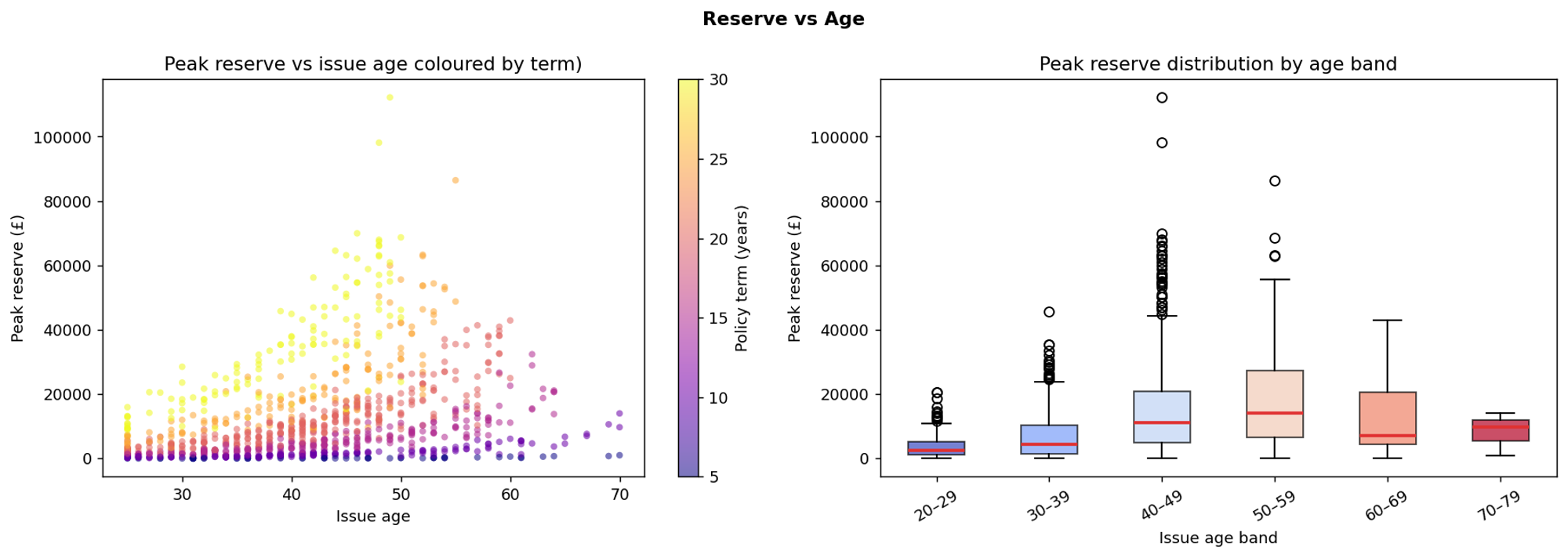}{0.90}
{Exploratory data analysis of the synthetic term-life portfolio showing peak reserve against issue age. The colour scale represents policy term, indicating how age and contract duration jointly affect reserve levels.}
{fig:reserve_vs_age}

\subsection{Feature Representation}

The final raw feature vector is:
\begin{equation}
x =
\left[
t,\;
a,\;
\rprice,\;
\rscen,\;
\frac{\premium}{\sumassured},\;
\sumassured,\;
\mortality(t)
\right].
\label{eq:feature_vector}
\end{equation}

Here, $t$ is elapsed duration, $a$ is issue age, $\rprice$ is the pricing interest rate, $\rscen$ is the scenario valuation interest rate, $\premium/\sumassured$ is the premium ratio, $\sumassured$ is sum assured, and $\mortality(t)$ is mortality intensity.

The separation between $\rprice$ and $\rscen$ is important. The pricing rate determines the premium at issue, while the scenario rate is used for reserve valuation under changed economic assumptions. Thus:
\begin{equation}
\premium = \premium(\rprice),
\qquad
r = \rscen.
\end{equation}

\subsection{Reserve-Ratio Target}

The model predicts a standardized reserve ratio rather than raw reserve. The reserve ratio is:
\begin{equation}
v(t)=\frac{\reserve(t)}{\sumassured}.
\end{equation}

The standardized target is:
\begin{equation}
z(t)=\frac{v(t)-\bar{v}}{\sigma_v}.
\end{equation}

The reserve is recovered by:
\begin{equation}
\hat{\reserve}(t)
=
\left(\hat{z}(t)\sigma_v+\bar{v}\right)\sumassured.
\end{equation}

The normalization parameters used in the runtime-analysis run are summarized in Table~\ref{tab:normalization}.

\begin{table}[H]
\centering
\caption{Feature and target normalization statistics.}
\label{tab:normalization}
\begin{tabular}{lc}
\toprule
Parameter & Value \\
\midrule
Interest-rate mean & 0.03992010 \\
Interest-rate standard deviation & 0.01476313 \\
Premium-ratio mean & 0.00470961 \\
Premium-ratio standard deviation & 0.00367158 \\
Target mean & 0.00800405 \\
Target standard deviation & 0.01264168 \\
\bottomrule
\end{tabular}
\end{table}

\section{Model Architecture and Learning Framework}

\subsection{Network Architecture and Training Configuration}

The reserve predictor is a fully connected neural network. The final model configuration is shown in Table~\ref{tab:model_config}.

\begin{table}[H]
\centering
\caption{Network architecture and training configuration.}
\label{tab:model_config}
\begin{tabular}{lc}
\toprule
Attribute & Value \\
\midrule
Input dimension & 7 \\
Hidden dimension & 256 \\
Number of layers & 6 \\
Activation function & $\tanh$ \\
Dropout & 0.0 \\
Skip connections & Enabled \\
Optimizer & Adam \\
Learning rate & 0.0002 \\
Weight decay & $5\times10^{-5}$ \\
Batch size & 64 \\
Epochs & 100 \\
Gradient clipping norm & 0.5 \\
Checkpoint frequency & Every 5 epochs \\
Early stopping patience & 45 epochs \\
Mixed precision & Disabled \\
\bottomrule
\end{tabular}
\end{table}

The $\tanh$ activation provides smooth derivatives, which is useful for PINN losses. Dropout is disabled because stochastic dropout can disturb derivative-based losses and sensitivity estimates.

\subsection{Physics-Informed Learning}

Unlike conventional neural networks, the proposed model is trained not only using reserve labels but also by enforcing compliance with Thiele's differential equation. During training, the predicted reserve is differentiated with respect to time, and the resulting derivative is compared with the right-hand side of Equation~\ref{eq:thiele}. The discrepancy forms the PDE residual loss, encouraging the network to learn reserve trajectories that satisfy the governing actuarial dynamics.

Consequently, the model is classified as a Physics-Informed Neural Network, where the governing actuarial equation forms an integral part of the optimization objective.

\subsection{Knowledge-Informed Learning}

In addition to the governing differential equation, the training process incorporates actuarial knowledge through domain-specific constraints. These include terminal boundary conditions, monotonic reserve behaviour with respect to key policy variables, reserve bounds, smooth reserve evolution, and scenario-based supervision under varying interest-rate environments. A summary of the knowledge-informed constraints is provided in Appendix~\ref{appendix:kinn}.

By combining supervised learning, physics-informed constraints, and actuarial knowledge, the proposed framework extends beyond a conventional PINN and is implemented as a hybrid Physics-Informed and Knowledge-Informed Neural Network.

\section{Loss Framework}

The total loss is a weighted sum of enabled losses:
\begin{equation}
L_{\mathrm{total}}
=
\sum_{k=1}^{K} w_k L_k.
\label{eq:total_loss}
\end{equation}

The main loss groups are shown in Figure~\ref{fig:loss_tree}. Detailed formulas for the individual losses are provided in Appendix~\ref{appendix:loss_formulas}.

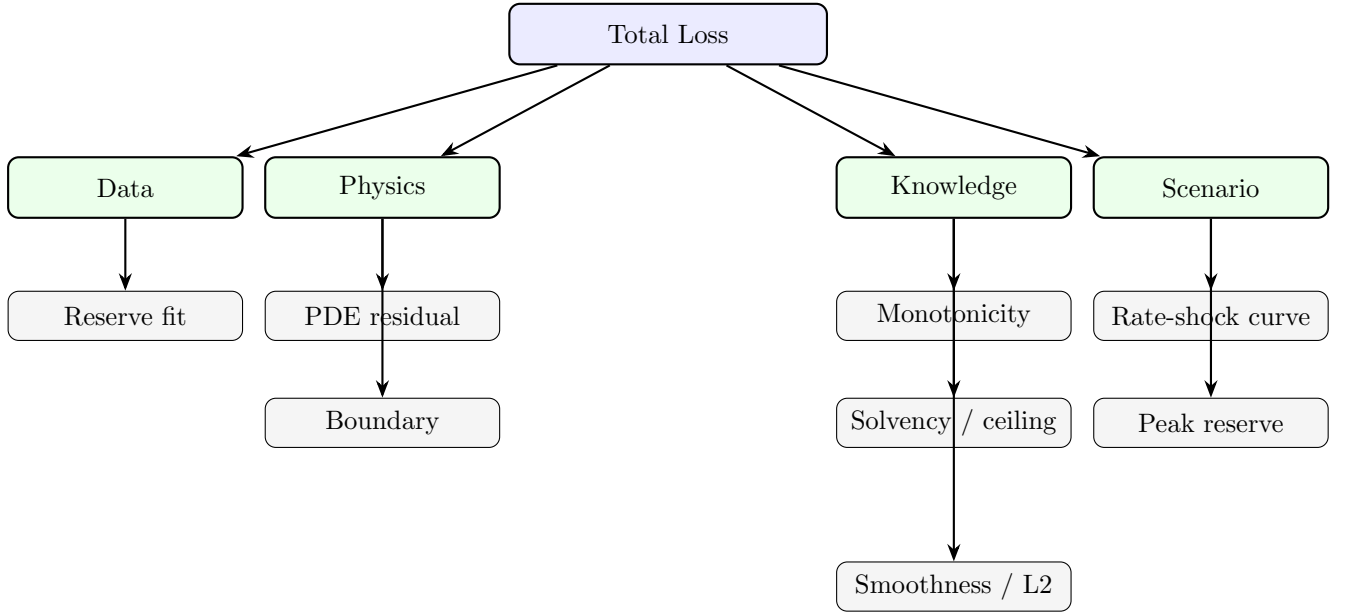
\begin{figure}[H]
\centering
\begin{tikzpicture}[
    >=Stealth,
    node distance=0.95cm,
    every node/.style={font=\small},
    root/.style={rectangle, rounded corners, draw=black, thick, fill=blue!8, minimum width=4.2cm, minimum height=0.8cm, align=center},
    group/.style={rectangle, rounded corners, draw=black, thick, fill=green!8, minimum width=3.1cm, minimum height=0.8cm, align=center},
    item/.style={rectangle, rounded corners, draw=black, fill=gray!8, minimum width=3.1cm, minimum height=0.65cm, align=center},
    arrow/.style={->, thick}
]

\node[root] (total) {Total Loss};

\node[group, below left=1.2cm and 3.5cm of total] (data) {Data};
\node[group, below left=1.2cm and 0.1cm of total] (physics) {Physics};
\node[group, below right=1.2cm and 0.1cm of total] (knowledge) {Knowledge};
\node[group, below right=1.2cm and 3.5cm of total] (scenario) {Scenario};

\node[item, below=of data] (data1) {Reserve fit};
\node[item, below=of physics] (physics1) {PDE residual};
\node[item, below=0.75cm of physics1] (physics2) {Boundary};
\node[item, below=of knowledge] (know1) {Monotonicity};
\node[item, below=0.75cm of know1] (know2) {Solvency / ceiling};
\node[item, below=1.5cm of know2] (know3) {Smoothness / L2};
\node[item, below=of scenario] (scen1) {Rate-shock curve};
\node[item, below=0.75cm of scen1] (scen2) {Peak reserve};

\draw[arrow] (total)--(data);
\draw[arrow] (total)--(physics);
\draw[arrow] (total)--(knowledge);
\draw[arrow] (total)--(scenario);

\draw[arrow] (data)--(data1);
\draw[arrow] (physics)--(physics1);
\draw[arrow] (physics)--(physics2);
\draw[arrow] (knowledge)--(know1);
\draw[arrow] (knowledge)--(know2);
\draw[arrow] (knowledge)--(know3);
\draw[arrow] (scenario)--(scen1);
\draw[arrow] (scenario)--(scen2);

\end{tikzpicture}
\caption{Loss framework used to train the PINN/KINN reserve model.}
\label{fig:loss_tree}
\end{figure}

The key scientific idea is that the model is not trained only for pointwise reserve accuracy. It is also trained to satisfy the reserve differential equation and actuarial behavioural constraints. This is what makes the model a PINN/KINN model rather than a standard neural network.

\section{Development Timeline}

The project was developed through a sequence of experimental runs. Runs 1--3 focused on basic stabilization. The initial model predicted values close to zero while target reserves were much larger, causing unstable training and large data loss. The main correction was to normalize the inputs and replace raw reserve targets with standardized reserve-ratio targets. Loss weights were also tuned, and reserve-trajectory comparison plots were added to monitor whether predicted curves matched classical actuarial behaviour.

Runs 4--8 focused on validation and physics refinement. A validation dashboard was added because reserve plots alone were insufficient. The PDE residual was reformulated in reserve-ratio space because raw-currency PDE residuals were numerically unstable. A sum-assured monotonicity loss and reserve dependency plots were also added to improve actuarial consistency and interpretability.

Runs 9--15 focused on sensitivity analysis and the premium-ratio refactor. Sensitivity and elasticity analysis were introduced to study how reserves respond to interest rate, mortality, premium, and sum assured. Premium was found to dominate the learned reserve response, so raw premium was replaced with premium ratio, $P/S$, across the dataset, model, losses, evaluator, and optimization modules. A fixed reference-policy setup was then used to produce clearer one-variable-at-a-time sensitivities.

Runs 16--21 focused on prototype optimization and interest-rate scenario behaviour. Optimization workflows were added for pricing-style, product-design-style, and portfolio-style experiments, showing that the trained reserve model could be used inside iterative decision loops. The final phase introduced pricing-rate and scenario-rate semantics, along with interest-rate scenario and peak-reserve losses, to improve reserve behaviour under shocked interest-rate conditions.

\section{Framework Architecture}

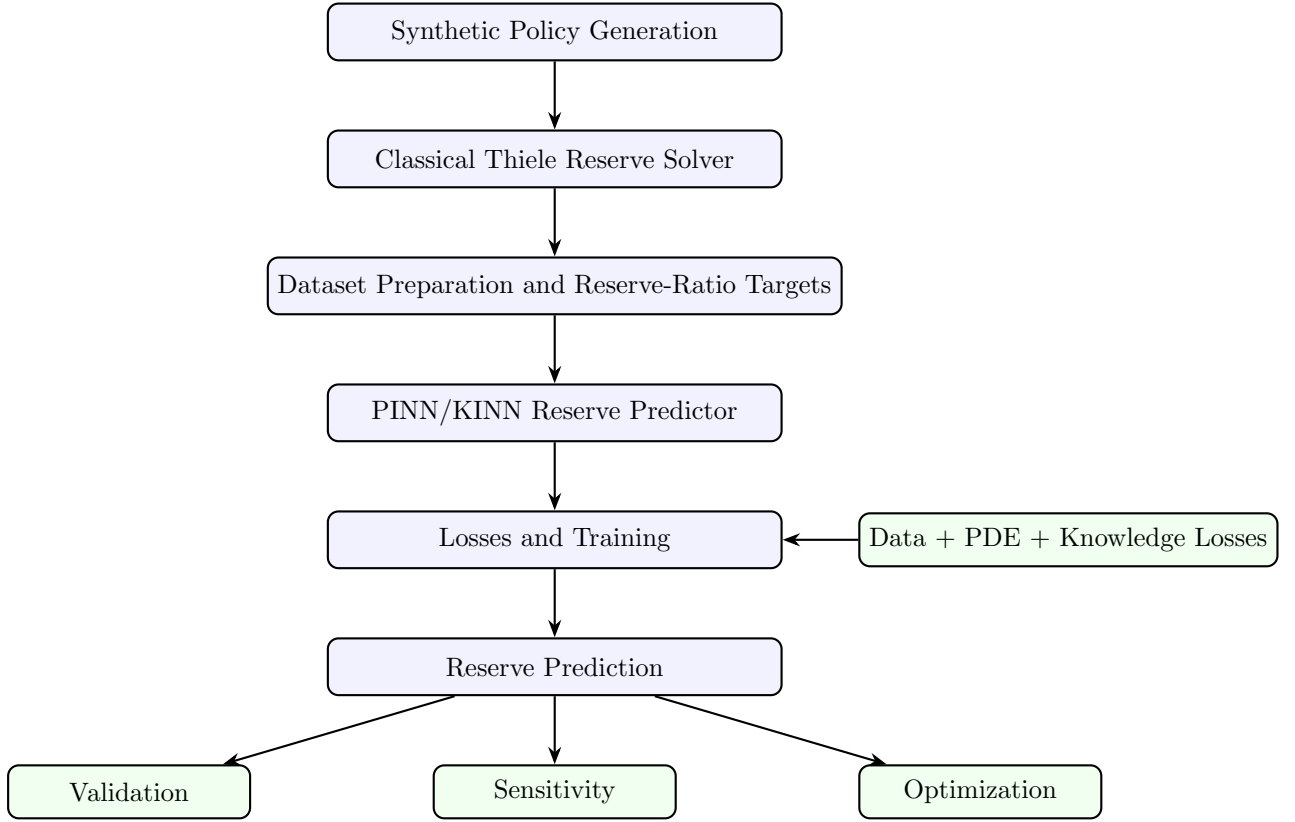
\begin{figure}[H]
\centering
\begin{tikzpicture}[
    >=Stealth,
    node distance=0.9cm,
    every node/.style={font=\small},
    block/.style={
        rectangle,
        rounded corners,
        draw=black,
        thick,
        minimum width=6.0cm,
        minimum height=0.75cm,
        align=center,
        fill=blue!5
    },
    side/.style={
        rectangle,
        rounded corners,
        draw=black,
        thick,
        minimum width=3.2cm,
        minimum height=0.7cm,
        align=center,
        fill=green!6
    },
    arrow/.style={->, thick}
]

\node[block] (policy) {Synthetic Policy Generation};
\node[block, below=of policy] (solver) {Classical Thiele Reserve Solver};
\node[block, below=of solver] (dataset) {Dataset Preparation and Reserve-Ratio Targets};
\node[block, below=of dataset] (model) {PINN/KINN Reserve Predictor};
\node[block, below=of model] (training) {Losses and Training};
\node[block, below=of training] (prediction) {Reserve Prediction};

\node[side, right=1.0cm of training] (losses) {Data + PDE + Knowledge Losses};
\node[side, below left=0.9cm and 1.0cm of prediction] (eval) {Validation};
\node[side, below=0.9cm of prediction] (sens) {Sensitivity};
\node[side, below right=0.9cm and 1.0cm of prediction] (opt) {Optimization};

\draw[arrow] (policy)--(solver);
\draw[arrow] (solver)--(dataset);
\draw[arrow] (dataset)--(model);
\draw[arrow] (model)--(training);
\draw[arrow] (losses)--(training);
\draw[arrow] (training)--(prediction);
\draw[arrow] (prediction)--(eval);
\draw[arrow] (prediction)--(sens);
\draw[arrow] (prediction)--(opt);

\end{tikzpicture}
\caption{Overall architecture of the Insurance Reserve Intelligence Platform.}
\label{fig:framework}
\end{figure}

The complete workflow is shown in Figure~\ref{fig:framework}. Compared with a simple prediction pipeline, the framework explicitly connects the PINN/KINN model to the loss system, training process, prediction workflow, and evaluation modules.

\section{Evaluation Methodology}

\subsection{Prediction Metrics}

The prediction metrics are:
\begin{align}
\mathrm{MSE} &= \frac{1}{N}\sum_{i=1}^{N}(V_i-\hat{V}_i)^2,\\
\mathrm{MAE} &= \frac{1}{N}\sum_{i=1}^{N}|V_i-\hat{V}_i|,\\
\mathrm{RMSE} &= \sqrt{\mathrm{MSE}},\\
R^2 &= 1-\frac{\sum_i(V_i-\hat{V}_i)^2}{\sum_i(V_i-\bar{V})^2}.
\end{align}

\subsection{Validation Diagnostics}

The validation framework evaluates accuracy, boundary satisfaction, PDE residual behaviour, monotonicity, and generalization. These diagnostics are necessary because high prediction accuracy does not guarantee actuarially correct behaviour under perturbations.

\subsection{Sensitivity and Elasticity}

Sensitivity analysis computes finite-difference reserve responses to one feature at a time:
\begin{equation}
\frac{\partial V}{\partial \rscen},\quad
\frac{\partial V}{\partial \mortality},\quad
\frac{\partial V}{\partial(P/S)},\quad
\frac{\partial V}{\partial S}.
\end{equation}

Elasticity converts sensitivities into comparable percentage effects:
\begin{equation}
E_x
=
\frac{\Delta V/V}{\Delta x/x}.
\end{equation}

\section{Results}

\subsection{Prediction Performance}

The final reported prediction performance is shown in Table~\ref{tab:performance}.

\begin{table}[H]
\centering
\caption{Final prediction performance on the test dataset.}
\label{tab:performance}
\begin{tabular}{lc}
\toprule
Metric & Value \\
\midrule
Mean Squared Error (MSE) & 1,470,771.88 \\
Mean Absolute Error (MAE) & 785.48 \\
Root Mean Squared Error (RMSE) & 1,212.76 \\
Coefficient of Determination ($R^2$) & 0.9887 \\
\bottomrule
\end{tabular}
\end{table}

These results indicate strong agreement with the classical actuarial solver at the aggregate test-set level.

\subsection{Reserve Trajectory Comparison}

Representative reserve curves show that the PINN/KINN model closely follows the classical solver for several test policies. The visible examples include mean errors of approximately 61, 153, 47, and 31 monetary units across four representative policies. The model captures the main reserve trajectory shape, although shorter-term and small-reserve policies remain more challenging.

\optionalfigure{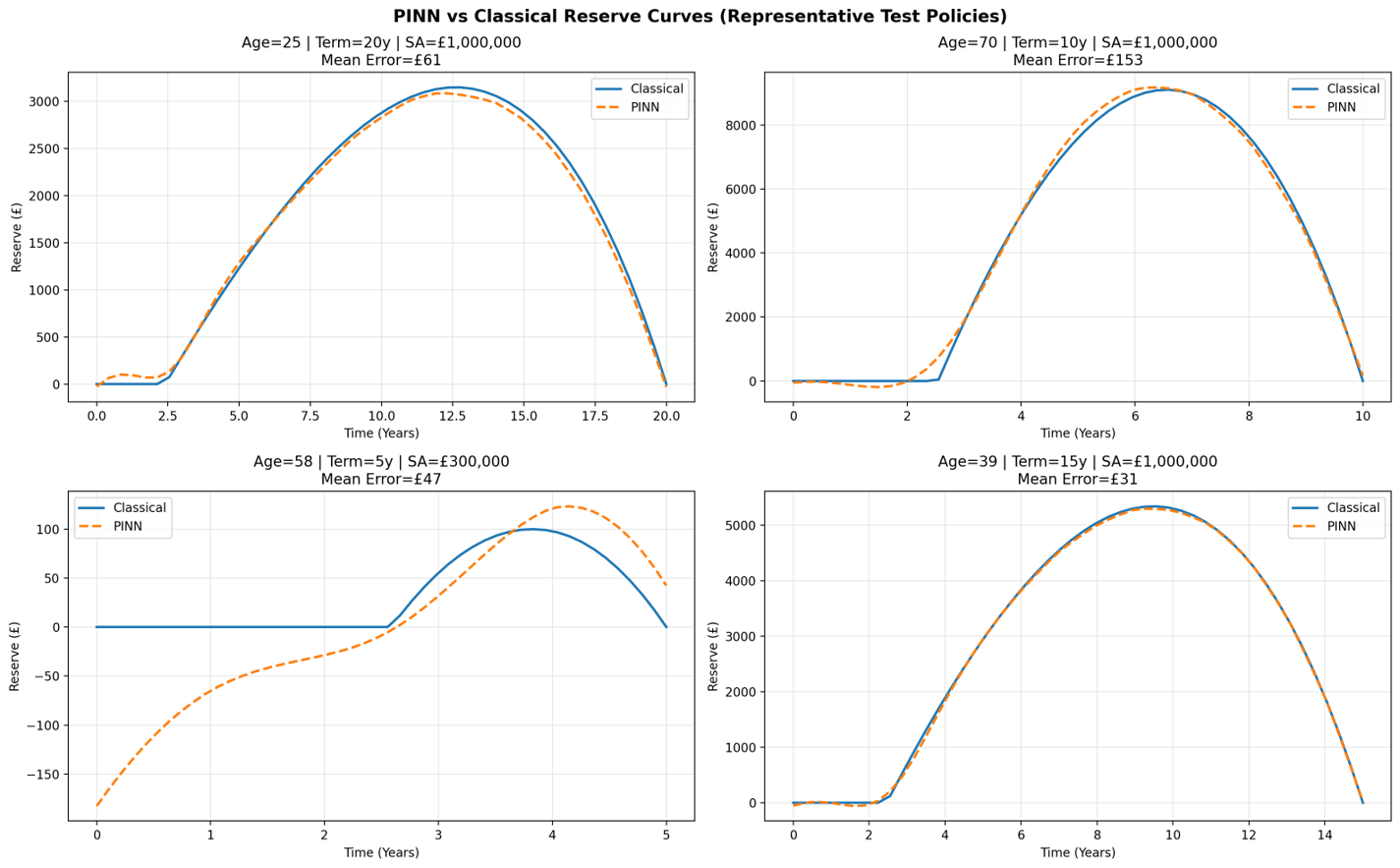}{0.82}
{Comparison between reserve trajectories predicted by the PINN/KINN model and the classical Thiele solver.}
{fig:reserve_prediction}

\subsection{Validation Results}

The validation dashboard indicates strong accuracy, physics, and boundary scores, but weak monotonicity and generalization. The visible validation results are summarized in Table~\ref{tab:validation_metrics}.

\begin{table}[H]
\centering
\caption{Validation diagnostic summary.}
\label{tab:validation_metrics}
\begin{tabular}{lc}
\toprule
Diagnostic & Value \\
\midrule
Accuracy score & 76 / 100 \\
Physics score & 99 / 100 \\
Boundary condition score & 89 / 100 \\
Actuarial monotonicity score & 27 / 100 \\
Generalization score & 0 / 100 \\
Mean relative error & 4.8\% \\
Mean boundary error $|V(T)|$ & approximately 111 \\
\bottomrule
\end{tabular}
\end{table}

These results show that the model is strong in aggregate fit and local physics behaviour, but not yet reliable for all actuarial monotonicity and out-of-distribution requirements.

\optionalfigure{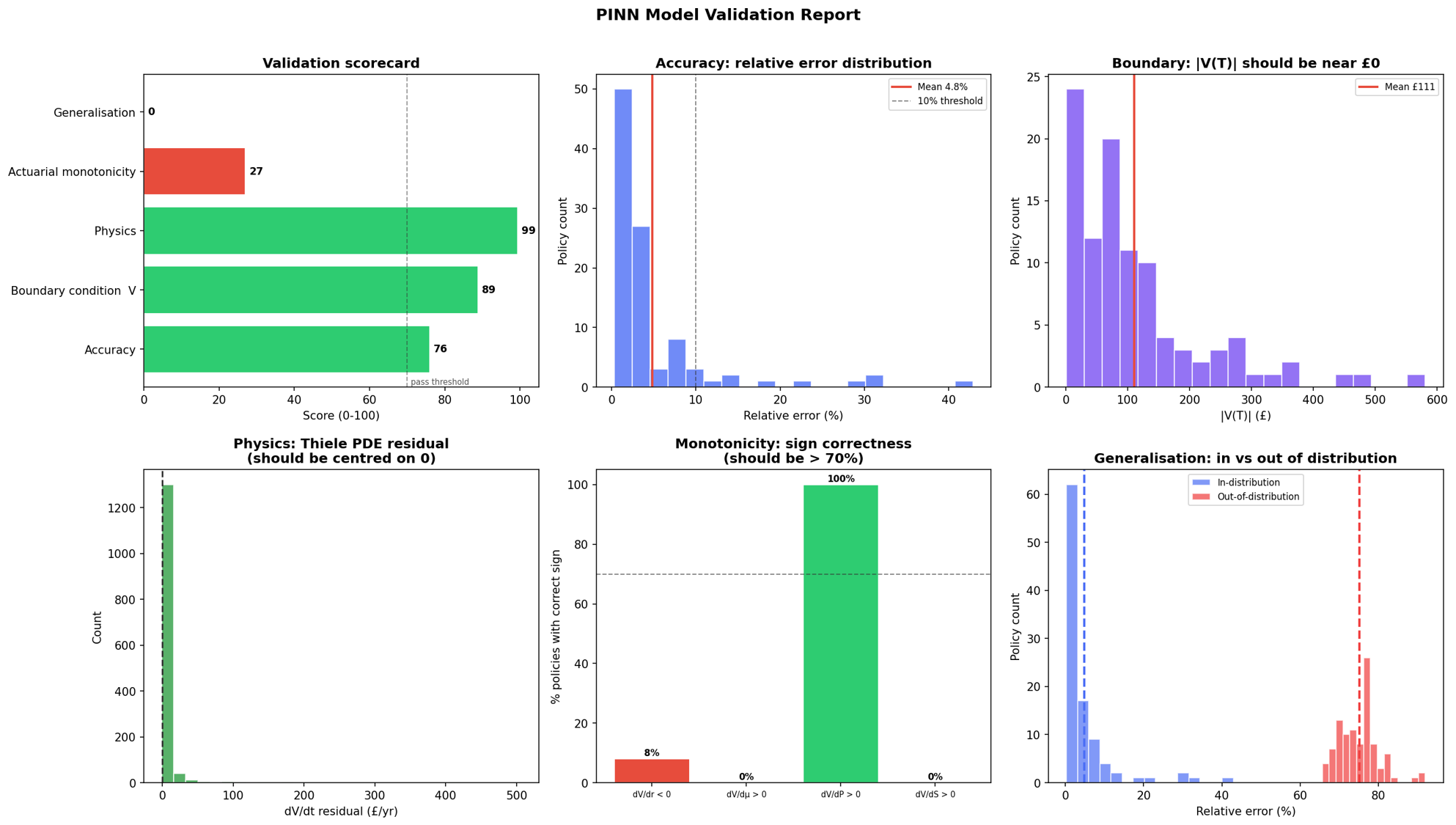}{0.95}
{Validation report summarizing accuracy, PDE residual behaviour, boundary error, monotonicity, and generalization.}
{fig:validation_report}

\subsection{Runtime Benchmark}

The runtime benchmark was executed on an Ubuntu-based Brev cloud environment with one NVIDIA T4 GPU with 16 GiB VRAM, 4 vCPUs, 16 GiB system RAM, and 256 GiB storage. The benchmark compares wall-clock reserve evaluation time for 200 policies using the classical Thiele solver and the trained PINN/KINN model. The reported speed-up is therefore specific to this hardware and software environment, but it provides a practical indication of the computational advantage of neural reserve inference. The result is shown in Table~\ref{tab:runtime}.

\begin{table}[H]
\centering
\caption{Runtime benchmark for reserve evaluation on 200 policies.}
\label{tab:runtime}
\begin{tabular}{lc}
\toprule
Metric & Value \\
\midrule
Hardware environment & NVIDIA T4 GPU, 4 vCPUs, 16 GiB RAM \\
Number of policies & 200 \\
Classical solver total time & 1.652123 s \\
PINN/KINN total time & 0.013822 s \\
Classical solver time per policy & 8.2606 ms \\
PINN/KINN time per policy & 0.0691 ms \\
Speed-up & 119.53x \\
\bottomrule
\end{tabular}
\end{table}

This result supports the main computational motivation for the neural surrogate. Once trained, the PINN/KINN model evaluates reserves substantially faster than repeated numerical solution of Thiele's equation. This speed advantage is especially relevant when reserve calculations are embedded inside sensitivity analysis, optimization loops, or scenario evaluation.

\subsection{Sensitivity Results}

The final sensitivity plot reports the raw partial derivative values shown in Table~\ref{tab:sensitivity_results}.

\begin{table}[H]
\centering
\caption{Reference-policy sensitivity results.}
\label{tab:sensitivity_results}
\begin{tabular}{lcc}
\toprule
Sensitivity & Value & Interpretation \\
\midrule
$dV/dr$ & -6,500.754 & Correct sign for interest-rate reserve response \\
$dV/d\mu$ & -817,044.188 & Incorrect sign for mortality response \\
$dV/d(P/S)$ & 225,128.984 & Correct positive premium-ratio response \\
$dV/dS$ & approximately 0.000 & Positive but very small response \\
\bottomrule
\end{tabular}
\end{table}

The most important remaining weakness is mortality sensitivity. The model still predicts a negative mortality response in the reference-policy sensitivity experiment, which is contrary to the expected actuarial direction for term-life reserves.

\optionalfigure{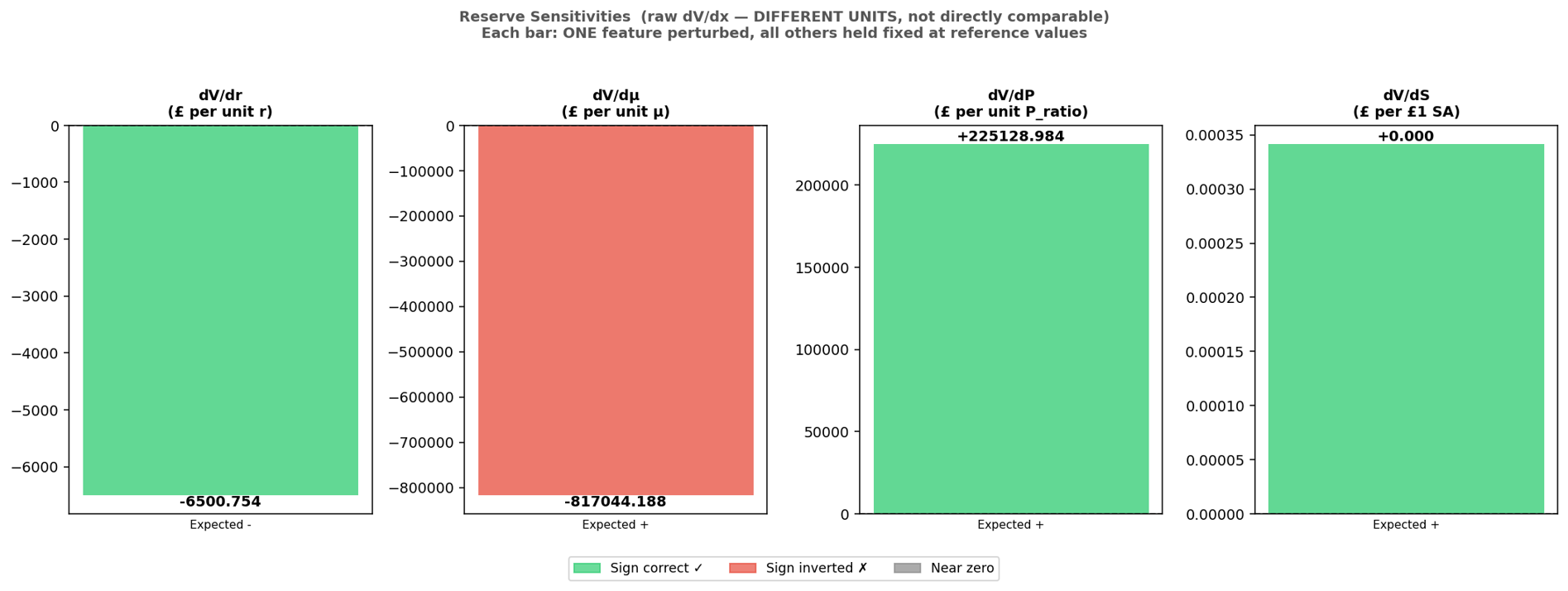}{0.82}
{Reference-policy sensitivity analysis using one-variable-at-a-time perturbations.}
{fig:sensitivity_report}

\subsection{Elasticity Results}

The elasticity plot expresses sensitivities as unit-free percentage responses. The visible elasticity values are summarized in Table~\ref{tab:elasticity_results}.

\begin{table}[H]
\centering
\caption{Reference-policy elasticity results.}
\label{tab:elasticity_results}
\begin{tabular}{lcc}
\toprule
Variable & Elasticity & Sign assessment \\
\midrule
Premium ratio & +267.92\% & Correct \\
Interest rate & -142.12\% & Correct \\
Mortality & -112.63\% & Incorrect \\
Sum assured & +76.03\% & Correct \\
\bottomrule
\end{tabular}
\end{table}

Elasticity analysis confirms that premium ratio has the strongest relative effect in the reference-policy experiment. It also confirms that mortality response remains the most important unresolved sensitivity issue.

\optionalfigure{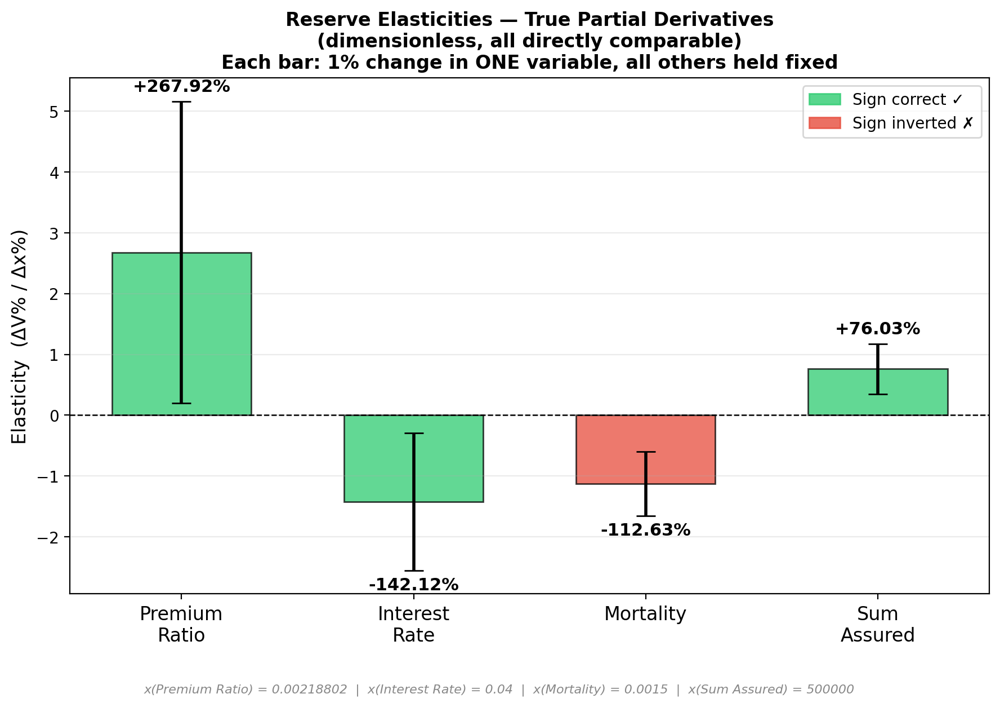}{0.82}
{Reserve elasticity analysis showing relative reserve response to one-percent changes in each feature.}
{fig:elasticity_report}

\subsection{Optimization Results}

The optimization plots show convergence of reserve values during pricing, product-design-style, and portfolio-style optimization experiments. The optimization objective and constraints differed across experiments; therefore, the reserve values are not directly comparable between optimization tasks.

\begin{table}[H]
\centering
\small
\caption{Optimization convergence observations from result plots.}
\label{tab:optimization_results}
\begin{tabular}{lccc}
\toprule
Experiment & Initial reserve & Final reserve & Observation \\
\midrule
Pricing optimization & approx 11,410 & approx 11,140 & Smooth convergence and stabilization \\
Product-design optimization & approx 11,430 & approx 10,720 & Stepwise convergence within early iterations \\
Portfolio optimization & approx 85,000 & approx 76,500 & Gradual portfolio-level reserve reduction \\
\bottomrule
\end{tabular}
\end{table}

The results demonstrate that the trained reserve model can be used inside optimization loops. However, stronger optimization claims would require objective values, constraints, success flags, and comparisons against classical solver-based optimization.

\section{Discussion}

The final model achieves strong aggregate prediction performance and closely follows the classical reserve solver for representative policies. The validation dashboard shows especially strong physics and boundary behaviour, suggesting that the model has learned a reserve surface that is broadly consistent with the classical equation. The runtime benchmark also supports the computational value of the surrogate, with approximately 119.53x faster reserve evaluation than the classical solver on the tested workload.

At the same time, the model is not fully resolved. The monotonicity score is low, and the generalization score is weak. Sensitivity analysis shows that mortality response remains incorrect in the reference-policy experiment. These findings show why actuarial neural models must be evaluated beyond aggregate prediction accuracy.

\section{Limitations}

The proposed framework has several limitations.

\begin{itemize}
    \item The model is developed and evaluated using synthetically generated term-life insurance data; validation on real-world insurance portfolios remains future work.
    \item The current implementation is limited to term-life insurance and does not yet support more complex insurance products.
    \item Although knowledge-informed constraints improve reserve behaviour, further work is required to enhance generalization and monotonicity across all policy variables.
    \item The optimization modules are proof-of-concept implementations, and the current evaluation does not include comprehensive baseline comparisons or ablation studies.
\end{itemize}

Despite these limitations, the proposed framework establishes a modular foundation for integrating classical actuarial modelling with physics-informed machine learning.

\section{Future Work}
\label{sec:future_work}

Future work should first address the scientific evaluation gaps. Baseline comparisons should be added against standard machine-learning models such as linear regression, random forest, XGBoost, and a standard multilayer perceptron. Ablation studies should also be conducted to measure the separate effects of the data loss, PDE residual loss, boundary loss, knowledge-informed actuarial losses, and interest-rate scenario losses. A broader runtime and scalability study should also be performed. The present benchmark was limited to 200 policies, where the classical solver required 1.652123 seconds and the PINN/KINN surrogate required 0.013822 seconds, giving a speed-up of 119.53$\times$. Future experiments should repeat this comparison on larger portfolios, such as 1,000, 10,000, and 50,000 policies, to evaluate whether the computational advantage remains stable at scale.

The framework should also be extended beyond synthetic term-life policies. Future versions should test real or industry-calibrated insurance datasets, support more complex insurance products, and include richer economic and actuarial scenarios involving lapse, inflation, stochastic mortality, and full interest-rate term structures. Additional uncertainty quantification methods, such as ensembles, quantile regression, conformal prediction, or Bayesian neural networks, may also be incorporated so that the reserve surrogate provides not only point estimates but also a measure of prediction uncertainty.

Finally, the long-term direction is to develop the platform into an insurance digital twin capable of continuous reserve monitoring, real-time scenario analysis, optimization, and AI-assisted actuarial decision support. This should be treated as a future system-level extension after the reserve surrogate has undergone stronger validation, broader benchmarking, and more formal model-governance review.

\section{Conclusion}

This paper presented an Insurance Reserve Intelligence Platform for term-life reserve modelling using a classical Thiele solver and a PINN/KINN neural reserve surrogate. The model is a PINN because it includes the residual of Thiele's differential equation in the loss function. It is a KINN because it also includes actuarial knowledge constraints such as boundary behaviour, monotonicity, reserve ceilings, solvency, smoothness, and scenario-specific losses.

The project developed a synthetic policy generation pipeline, reserve-ratio dataset formulation, configurable multi-loss training framework, validation diagnostics, sensitivity analysis, elasticity analysis, runtime benchmarking, and prototype optimization workflows.

The final model achieved strong aggregate prediction performance, with $R^2=0.9887$, MAE 785.48, and RMSE 1,212.76. It also achieved a 119.53x runtime speed-up over the classical solver in the measured benchmark. At the same time, monotonicity and generalization remain important limitations. The main conclusion is that physics- and knowledge-informed neural networks can approximate classical reserve calculations effectively, but actuarial reliability requires evaluation of physics consistency, boundary behaviour, monotonicity, scenario response, sensitivity behaviour, optimization stability, runtime performance, and generalization.

\appendix

\section{Variable Definitions}

\begin{table}[H]
\centering
\caption{Main variables used in the paper.}
\begin{tabular}{ll}
\toprule
Symbol & Meaning \\
\midrule
$t$ & Elapsed policy duration \\
$T$ & Policy term or maturity time \\
$a$ & Issue age \\
$\reserve(t)$ & Reserve at time $t$ \\
$\premium$ & Premium \\
$\sumassured$ & Sum assured or death benefit \\
$\mortality(t)$ & Mortality intensity at time $t$ \\
$r$ & Generic valuation interest rate \\
$\rprice$ & Pricing interest rate used to calculate premium \\
$\rscen$ & Scenario interest rate used for reserve valuation \\
$v(t)$ & Reserve ratio, $\reserve(t)/\sumassured$ \\
$z(t)$ & Standardized reserve-ratio target \\
$\Delta r$ & Interest-rate shock size \\
\bottomrule
\end{tabular}
\end{table}

\section{Detailed Loss Formulas}
\label{appendix:loss_formulas}

The supervised data loss is:
\begin{equation}
L_{\mathrm{data}}
=
\frac{1}{N}
\sum_{i=1}^{N}
(\hat{z}_i-z_i)^2.
\label{eq:data_loss}
\end{equation}

The reserve-ratio version of Thiele's equation is:
\begin{equation}
\frac{d\hat{v}}{dt}
=
\rscen \hat{v}
+
\frac{\premium}{\sumassured}
-
\mortality(t)(1-\hat{v}).
\label{eq:ratio_thiele}
\end{equation}

The PDE loss is:
\begin{equation}
L_{\mathrm{PDE}}
=
\frac{1}{N}
\sum_{i=1}^{N}
\left[
\frac{d\hat{v}_i}{dt}
-
\left(
\rscen \hat{v}_i
+
\frac{\premium_i}{\sumassured_i}
-
\mortality_i(1-\hat{v}_i)
\right)
\right]^2.
\label{eq:pde_loss}
\end{equation}

The boundary loss is:
\begin{equation}
L_{\mathrm{boundary}}
=
\frac{1}{N_T}
\sum_{i=1}^{N_T}
\hat{\reserve}_i(T)^2.
\label{eq:boundary_loss}
\end{equation}

The mortality monotonicity loss is:
\begin{equation}
L_{\mu}
=
\frac{1}{N}
\sum_{i=1}^{N}
\max\left(0,-\frac{\partial \hat{\reserve}_i}{\partial \mortality}\right).
\label{eq:mortality_loss}
\end{equation}

The interest-rate sensitivity loss is:
\begin{equation}
L_{r}
=
\frac{1}{N}
\sum_{i=1}^{N}
\ell_{\delta}
\left(
\frac{\partial \hat{v}_i}{\partial \rscen},
s_i^r
\right).
\label{eq:interest_sensitivity_loss}
\end{equation}

The solvency loss is:
\begin{equation}
L_{\mathrm{solvency}}
=
\frac{1}{N}
\sum_{i=1}^{N}
\max(0,-\hat{\reserve}_i).
\label{eq:solvency_loss}
\end{equation}

The reserve ceiling loss is:
\begin{equation}
L_{\mathrm{ceiling}}
=
\frac{1}{N}
\sum_{i=1}^{N}
\max(0,\hat{\reserve}_i-\sumassured_i).
\label{eq:ceiling_loss}
\end{equation}

The smoothness loss is:
\begin{equation}
L_{\mathrm{smooth}}
=
\frac{1}{N}
\sum_{i=1}^{N}
\left(
\frac{\partial^2 \hat{\reserve}_i}{\partial t^2}
\right)^2.
\label{eq:smoothness_loss}
\end{equation}

The sum-assured monotonicity loss is:
\begin{equation}
L_{S}
=
\frac{1}{N}
\sum_{i=1}^{N}
\max\left(0,-\frac{\partial \hat{\reserve}_i}{\partial \sumassured}\right).
\label{eq:sum_assured_loss}
\end{equation}

The interest-rate scenario loss is:
\begin{equation}
L_{\mathrm{scenario}}
=
\frac{1}{2}
\ell_{\delta}
\left(
\hat{v}_{r-\Delta r},
v_{r-\Delta r}
\right)
+
\frac{1}{2}
\ell_{\delta}
\left(
\hat{v}_{r+\Delta r},
v_{r+\Delta r}
\right).
\label{eq:scenario_loss}
\end{equation}

The peak-reserve loss is:
\begin{equation}
L_{\mathrm{peak}}
=
\frac{1}{2}
\ell_{\delta}
\left(
\hat{v}^{\max}_{r-\Delta r},
v^{\max}_{r-\Delta r}
\right)
+
\frac{1}{2}
\ell_{\delta}
\left(
\hat{v}^{\max}_{r+\Delta r},
v^{\max}_{r+\Delta r}
\right).
\label{eq:peak_loss}
\end{equation}

The L2 regularization loss is:
\begin{equation}
L_{\mathrm{L2}}
=
\sum_j
\|\theta_j\|_2^2.
\label{eq:l2_loss}
\end{equation}

The Huber loss is:
\begin{equation}
\ell_{\delta}(a,b)
=
\begin{cases}
\frac{1}{2}(a-b)^2, & |a-b|\leq \delta,\\
\delta\left(|a-b|-\frac{1}{2}\delta\right), & |a-b|>\delta.
\end{cases}
\label{eq:huber_loss}
\end{equation}

\section{Knowledge-Informed Constraints}
\label{appendix:kinn}

\begin{table}[H]
\centering
\caption{Knowledge-informed constraints incorporated during training.}
\label{tab:kinn_constraints}
\begin{tabular}{p{4cm}p{9cm}}
\toprule
\textbf{Constraint} & \textbf{Purpose} \\
\midrule
Boundary condition &
Enforces the terminal reserve condition $V(T)=0$. \\
Monotonicity &
Encourages actuarially consistent reserve behaviour with respect to mortality, premium, interest rate, and sum assured. \\
Reserve positivity &
Discourages strongly negative reserve values under normal policy conditions. \\
Reserve ceiling &
Prevents reserves from exceeding reasonable actuarial limits without justification. \\
Smoothness &
Reduces unrealistic oscillations in reserve trajectories. \\
Scenario constraints &
Improves reserve behaviour under varying interest-rate environments using scenario and peak losses. \\
\bottomrule
\end{tabular}
\end{table}

\section{Development Summary}

\begin{table}[H]
\centering
\caption{Summary of major development stages.}
\begin{tabular}{p{0.18\textwidth}p{0.34\textwidth}p{0.34\textwidth}}
\toprule
Stage & Problem & Main Fix \\
\midrule
Runs 1--3 & Raw reserves caused unstable training & Standardized reserve-ratio target \\
Runs 4--8 & No systematic validation and unstable PDE loss & Added validation dashboard and reformulated PDE in reserve-ratio space \\
Runs 9--15 & Sensitivities were difficult to interpret and premium dominated learning & Added sensitivity/elasticity analysis and replaced premium with premium ratio \\
Runs 16--21 & Need optimization and stronger interest-rate scenario behaviour & Added prototype optimization workflows, pricing/scenario rate semantics, scenario loss, and peak-reserve loss \\
\bottomrule
\end{tabular}
\end{table}

\end{document}